\documentclass[conference]{IEEEtran}
\IEEEoverridecommandlockouts
\usepackage{cite}
\usepackage{amsmath,amssymb,amsfonts}
\usepackage{graphicx}
\usepackage{textcomp}
\usepackage{xcolor}
\usepackage[shortcuts,acronym]{glossaries}
\usepackage{multicol} 
\usepackage{threeparttable}
\usepackage{multirow}
\usepackage{subfigure}
\usepackage{cancel}
\usepackage{algorithm}
\usepackage{algpseudocode}
\usepackage[capitalize]{cleveref}
\usepackage{bm}

\newacronym{ofdm}{OFDM}{Orthogonal Frequency-Division Multiplexing}
\newacronym{ngp}{Instant-NGP}{Instant Neural Graphics Primitives}
\newacronym{nerf}{NeRF}{Neural Radiance Field}
\newacronym{cv}{CV}{computer vision}
\newacronym{psnr}{PSNR}{peak signal-to-noise ratio}
\newacronym{ssim}{SSIM}{structural similarity index measure}
\newacronym{spp}{SPP}{switched Poisson process}
\newacronym{ugv}{UGV}{unmanned ground vehicle}
\newacronym{uav}{UAV}{unmanned aerial vehicle}
\newacronym{bs}{BS}{base station}
\newacronym{lpips}{LPIPS}{learned perceptual image patch similarity}
\newacronym{mse}{MSE}{mean squared error}
\newacronym{cuda}{CUDA}{Compute Unified Architecture}
\newacronym{ge}{G-E}{Gilbert-Elliot}
\newacronym{mmpp}{MMPP}{Markov-modulated Poisson Process}
\newacronym{qos}{QoS}{Quality of Service}
\newacronym{aoi}{AoI}{Age of Information}
\newacronym{drl}{DRL}{Deep Reinforcement Learning}
\newacronym{voi}{VoI}{Value of Information}
\newacronym{ros}{ROS}{Robot Operating System}
\newacronym{ppo}{PPO}{Proximal Policy Optimization}
\newacronym{bleu}{BLEU}{Bilingual Evaluation Understudy}
\newacronym{mat}{MAT}{Maximum AoI Threshold}
\newacronym{mlp}{MLP}{multilayer perceptron}

\def\BibTeX{{\rm B\kern-.05em{\sc i\kern-.025em b}\kern-.08em
    T\kern-.1667em\lower.7ex\hbox{E}\kern-.125emX}}
\begin{document}

\title{Timeliness-Fidelity Tradeoff in 3D Scene Representations\\

}

\author{
  \IEEEauthorblockN{Xiangmin Xu\textsuperscript{1},
                    Zhen Meng\textsuperscript{1},
                    Yichi Zhang\textsuperscript{1},
                    Changyang She\textsuperscript{2},
                    Philip G. Zhao\textsuperscript{3}}
\textit{\textsuperscript{1}School of Engineering, University of Glasgow, UK} \\
\textit{\textsuperscript{2}School of Electrical and Information Engineering, University of Sydney, Australia.} \\
\textit{\textsuperscript{3}Department of Computer Science, University of Manchester, UK.} \\
\textit{x.xu.1@research.gla.ac.uk, z.meng.1@research.gla.ac.uk, y.zhang.13@research.gla.ac.uk,} \\
\textit{shechangyang@gmail.com, philip.zhao@manchester.ac.uk}
}


\maketitle
 
\begin{abstract}
Real-time three-dimensional (3D) scene representations serve as one of the building blocks that bolster various innovative applications, e.g., digital manufacturing, Virtual/Augmented/Extended/Mixed Reality (VR/AR/XR/MR), and the metaverse. Despite substantial efforts that have been made to real-time communications and computing, real-time 3D scene representations remain a challenging task. This paper investigates the tradeoff between timeliness and fidelity in real-time 3D scene representations. Specifically, we establish a framework to evaluate the impact of communication delay on the tradeoff, where the real-world scenario is monitored by multiple cameras that communicate with an edge server. To improve fidelity for 3D scene representations, we propose to use a single-step Proximal Policy Optimization (PPO) method that leverages the Age of Information (AoI) to decide if the received image needs to be involved in 3D scene representations and rendering. We test our framework and the proposed approach with different well-known 3D scene representation methods. Simulation results reveal that real-time 3D scene representation can be sensitively affected by communication delay, and our proposed method can achieve optimal 3D scene representation results.






\end{abstract}


\section{Introduction}
3D scene representations involve the process of capturing, interpreting, and recreating a three-dimensional representation of an object or a scene from two-dimensional images or sensor data~\cite{tewari2022advances}. By enabling comprehensive understanding and interaction with the visual world, it served as the fundamental building blocks for various emerging applications, e.g., digital manufacturing~\cite{industrialnerf}, autonomous driving~\cite{feifeidriving}, Virtual/Augmented/Extended/Mixed Reality (VR/AR/XR/MR)~\cite{aoipetar}, and the metaverse~\cite{zhenjsac}. \glspl{nerf}~\cite{nerf}, a promising 3D scene representations approach, leverages a neural network to implicitly infer scene radiance and density, capturing detailed perceptional geometry and illumination in the 3D scene representations. Recent results show that it can render 2D photos into 3D scenes in a few milliseconds~\cite{instantngp}. However, most of the existing \gls{nerf} algorithms require a considerable amount of computation resources, and the processing time remains a challenging issue of real-time interactions. 

Furthermore, the communication delay may become the bottleneck in real-time interactions.

\begin{figure}
    \centering
    \includegraphics[width=\linewidth]{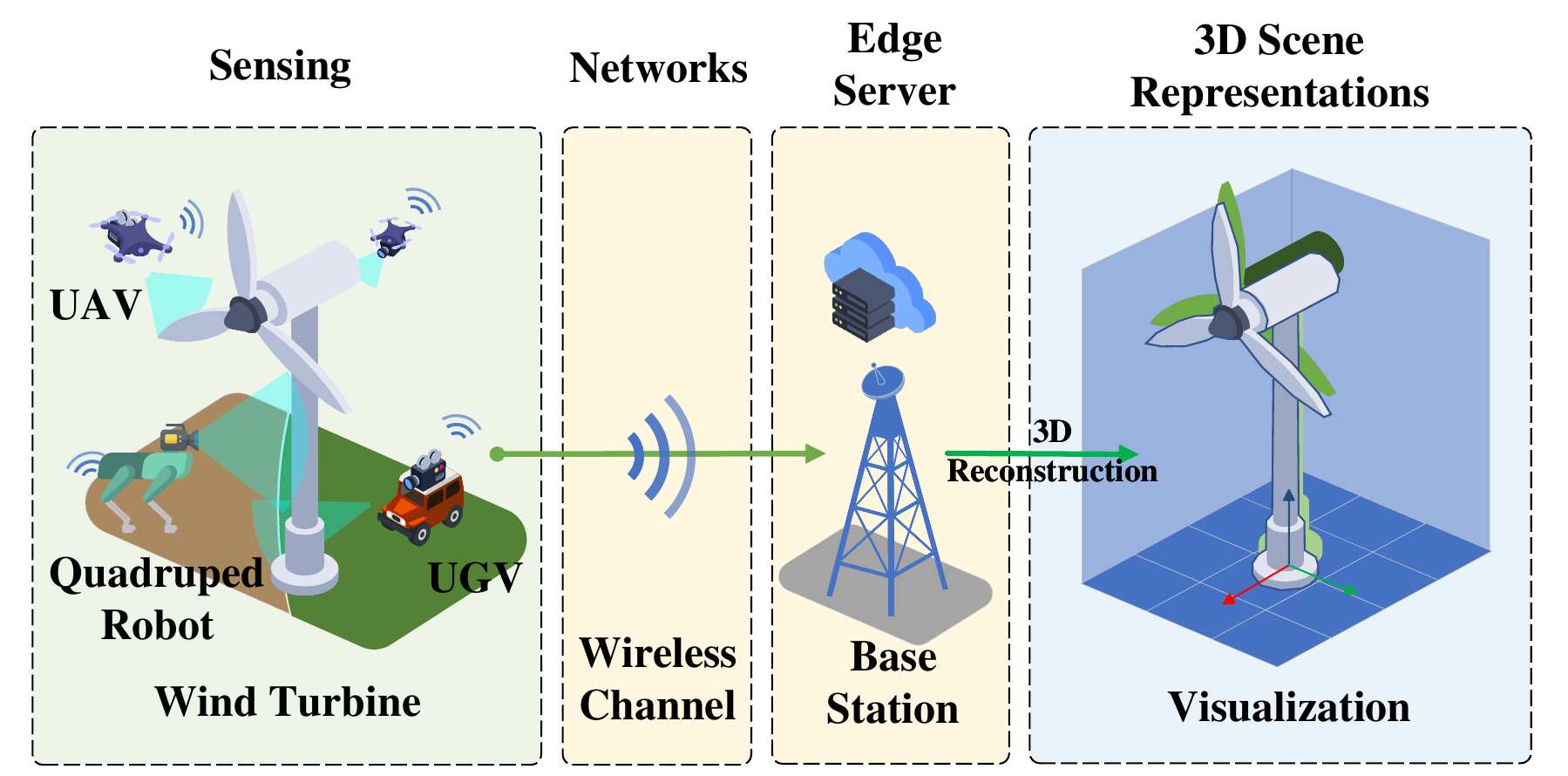}
    \caption{System model.}
    \label{fig: poster}
\end{figure}
In the 5G New Radio, a variety of communication KPIs have been specified for different application scenarios~\cite{3GPP}. In particular,  the typical KPIs in the most related scenario, i.e., video streaming include 1 Gbps (smooth play) or 2.35 Gbps (interactive) data rate to support 10~ms latency for 8K resolution, 120~FPS, and 360~degree visual field VR video streaming~\cite{mangiante2017vr}. It is worth noting that there is a gap between these communication KPIs and the requirements of real-time 3D scene representations. This gap leads to unprecedented challenges and hinders the emergence of new 3D scene representation-related applications. 

The overall system performance of real-time 3D scene representations depends on numerous factors beyond communications, such as sensing, sampling, computing (inference and rendering), storage, etc. This motivates us to investigate to adopt interdisciplinary approaches to address these challenges, where the distinct characteristics of real-time 3D scene representation-related applications should be considered. 

3D scene representations also pose notable challenges to timeliness and fidelity~\cite{tewari2022advances}. For example, as shown in Fig.~\ref{fig: poster}, different autonomous robots take 2D images of a target scenario and transmit the images to an edge server for constructing the 3D scene in real time. Due to communication delay and jitter, the images may not arrive at the edge server simultaneously.
If the edge server only waits for the first few images for 3D scene representations, the timeliness is guaranteed, but the fidelity could be poor since some images experiencing long latency are not exploited in the 3D scene representations. On the other hand, data fidelity can be improved by waiting for more images, but this leads to the deterioration of timeliness.

To investigate the timeliness-fidelity tradeoff in 3D scene representations, we establish a framework to evaluate the impact of communication delay on the tradeoff, where the real-world scenario is monitored by multiple cameras that communicate with an edge server. To achieve optimal fidelity 3D scene representation results, we propose to use single-step \gls{ppo} method that leverages the \gls{aoi} to choose the received images for 3D scene representations and rendering. We test our framework and the proposed approach with different well-known 3D scene representation methods. Simulation results reveal that the quality of 3D scene representations highly depends on the communication delay, and our proposed method can achieve optimal 3D scene representation results. To the best of our knowledge, this is the first paper investigating the impact of communication delay on real-time 3D scene representations.
\section{Related Work}
Significant contributions have been made in the existing literature for enhancing the inference time and fidelity of \glspl{nerf}~\cite{rosinol2022nerf,Yu2022MonoSDF,yu2023nerfbridge,Zhu2022CVPR,Sucar:etal:ICCV2021}. The authors in~\cite{rosinol2022nerf} leveraged dense monocular SLAM to deliver precise pose estimation and depth maps, optimizing the fitting of a neural radiance field to the scene in real time. The authors in~\cite{Yu2022MonoSDF} showed that the prediction of depth and normal cues generated by the general monocular estimators can significantly improve the convergence speed and quality of the radiance field scene representation. Nerf-Bridge proposed in~\cite{yu2023nerfbridge} facilitates the rapid advancement of research on applications of \glspl{nerf} in robotics by providing an extensible interface between Nerfstudio~\cite{nerfstudio} and the \gls{ros} to the efficient training pipelines. The contributions of these works are limited to the field of \gls{cv}, however, the communications of \gls{uav}, \gls{ugv}, and quadruped robotics are very vulnerable to interferences in wireless communications, showing the necessity of investigating the impacts of communication systems on real-time 3D scene representations. 

On the other hand, different communication metrics that take the content of the message into account, such as the \gls{aoi}, \gls{voi}, and semantic metric, e.g.,~\gls{bleu}~\cite{bleu}, ~\gls{ssim}~\cite{ssim} have been considered in different tasks including \gls{ugv} control~\cite{10013736}, image transmission and~\cite{9959884} and image classification~\cite{9606667}. The author in~\cite{agheli2023effective} introduced a pull-based communication system, where a sensing agent adjusts an actuation agent through a query control policy. The results show that the proposed agent is able to respond to changing information sources and is updated to be effective in achieving specific goals.
The authors in~\cite{8262774} explored different policies that either involve waiting for complete acknowledgments or partial acknowledgments to enhance \gls{aoi} performance in a broadcast network. 
The authors in~\cite{9322193} introduced an optimal scheduling policy based on thresholds to centrally manage the timeliness and energy cost of updates. 
The authors in~\cite{9467360} investigated a tradeoff between the controller’s sampling preference and the communication loads among devices in a software-defined Internet-of-Things networking system. 
The work most relevant to us is~\cite{chen2023improving}, where the authors designed a fusion center that optimizes data timeliness and fidelity by waiting for observations from all nodes and partial nodes. The authors used the \gls{aoi} for timeliness and the \gls{mse} for fidelity evaluation. These works provided useful insights on some fundamental tradeoffs, but did not consider 3D scene representations. 

\section{System Model}
 \begin{figure}
    \centering
    \includegraphics[width=0.39\textwidth]{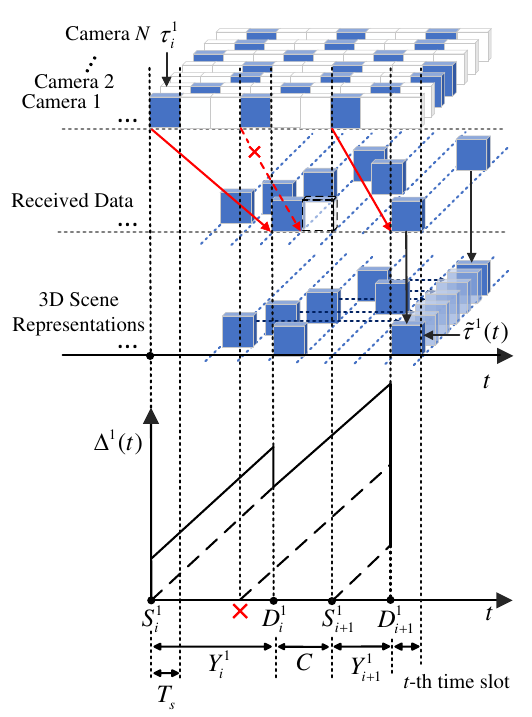}
    \caption{Time Sequence Diagram.}
    \label{fig: time}
\end{figure}
\subsection{Overview} 
As shown in Fig~\ref{fig: poster}, we propose a sensing, communications, and 3D scene representations framework. First, a team of autonomous robots (e.g., \glspl{uav}, \glspl{ugv}, and quadruped robots, etc.) take 2D images of moving objects (e.g., wind turbine) that need to be rendered in 3D in different angle-views. Images and poses taken from different robotics are transmitted to an edge server via a wireless channel. The 3D scene representations are performed at the edge server.

The time sequence diagram of the framework is shown in Fig.~\ref{fig: time}. Time is discretized into slots, and the duration of each time slot is denoted by $T_s$. In the framework, $N$ cameras indexed by $n = \{1,...,N\}$ generate images every $C$ time slot and transmit them to the edge server. The pose of the $n$-th camera denoted by 
$\mathbf{p}^n(t) = [x^n(t), y^n(t), z^n(t), \theta^n(t), \phi^n(t)] \in \mathbb{R}^{1\times5}$
, where $[x^n(t), y^n(t), z^n(t)]$ is the 3D location and $[\theta^n(t), \phi^n(t)]$ is the 2D viewing direction. We assume the poses of the cameras alter as the robots move, and are available to the edge server.

We define the $i$-th image of the $n$-th camera by ${\tau}_i^n, i=1,2,...I$. The $n$-th camera starts to transmit ${\tau}_i^n$ at time slot $S_i^n$, and the delivery is finished at time slot $D^n_i$. The corresponding transmission duration is $Y^n_i =D^n_i- S_i^n$, which depends on the wireless communication channel. We consider a single-cell \gls{ofdm} uplink wireless network system. Specifically, there are $M$ orthogonal subchannels assigned to the $N, N=M$ cameras. Due to the channel fading, the transmission time is stochastic. Thus, we assume that the transmission duration of an image may follow arbitrary distributions.
In addition, if the last image is still in transmission, the new image will not be transmitted to the edge server. In this case, we can model the queue system at the camera as a D/G/1/0 model.

In the $t$-th time slot, the most recently received image of $n$-th camera is generated at time
\begin{align}
U^n(t) = \text{max}\{S^n_i : D^n_i \leq t\}.    
\end{align}
The \gls{aoi} of the $n$-th camera is defined as~\cite{6195689},
\begin{align}
\Delta^n(t)= t - U^n(t). 
\end{align}
We denote the latest images updated from the $N$ cameras by $\widetilde{\mathcal{T}}(t) =[\widetilde{\tau}^1(t), \widetilde{\tau}^2(t),...,\widetilde{\tau}^n(t),...,\widetilde{\tau}^N(t)]$.
Based on their \gls{aoi}s, we design an agent for scene representation at the edge server, where the indicator, $\omega^n(t)$, represents whether the latest image from the $n$-th camera $\widetilde{\tau}^n(t)$ is used for 3D scene representations in the $t$-th time slot, $n=1,2,...,N$. If $\widetilde{\tau}^n(t)$ is not used for 3D scene representations, then $\omega^n(t) = 0$. Otherwise, $\omega^n(t) = 1$. The decision of the agent in the $t$-th time slot is denoted by $\Omega (t) = [\omega^1(t), \omega^2(t),...,\omega^N(t)]$. We denote the set of images used for 3D scene representations in the $t$-th time slot by $\mathbb{D}(t) =\{\widetilde{\tau}^n(t) \mid \omega^n(t) = 1,\ n=1,...,N\}$ and the corresponding set of poses is denoted by $\mathbb{P}(t) =\{\mathbf{p}^n(t) \mid \omega^n(t) = 1,\ n=1,...,N\}$.

\subsection{3D Scene Representations}
For 3D scene representations, we use the \gls{nerf} method~\cite{nerf}, which models the continuous radiance fields of a scene with a \gls{mlp}. Specifically, in the $t$-th time slot, the \gls{mlp} denoted by $\mathcal{F}_\Theta$ takes poses $\mathbb{P}(t)$ as input and outputs the volume density $\sigma(t)$ and view-dependent RGB color $\textbf{c}(t) = [r, g, b]$, which is expressed by 
\begin{equation}
\{\textbf{c}(t), \sigma(t)\} = {\mathcal{F}_\Theta}\left({\mathbb{P}(t), {\alpha _\Theta }} \right),
\label{eqn: train}
\end{equation}

where the parameters of the neural network are denoted by ${\alpha_\Theta}$. 

To visualize the 3D scene representations, a 2D image $\mathbf{\hat{I}}(t)$ can be obtained by volume rendering~\cite{volumerendering}, which is expressed by
\begin{equation}
{\mathbf{\hat{I}}}(t) = {{\cal F}_r}\left( {\textbf{c}(t), \sigma(t),\mathbf{p}_v,\alpha_r} \right),
\label{eqn:render}
\end{equation}
where $\mathbf{p}_v = [x_v, y_v, z_v, \theta_v, \phi_v] \in \mathbb{R}^{1\times5}$ is the pose of the desired view, consisting of the 3D location $[x_v, y_v, z_v]$ and the 2D viewing direction $[\theta_v, \phi_v]$. The parameters of the rendering function are denoted by ${\alpha _r}$. Specifically, for the function of volume rendering ${\cal F}_r(\cdot)$, given a ray $\mathbf{r}(q) = \mathbf{o} + q\mathbf{d}, \left|\mathbf{d}\right| = 1$ emanating from the camera position $\mathbf{o}$ and direction $\mathbf{d}$ defined by the specified camera pose and intrinsic parameters.
 
The rendered color of each pixel corresponding to this ray can be calculated by integrating the cumulative opacity-weighted radiance, which is expressed by
\begin{equation}
\label{Cr}
        C(\textbf{r}) = \int_{q_n}^{q_f} T(q)\sigma(\textbf{r}(q))\textbf{c}\left(\textbf{r}(q), \textbf{d}\right)dq,
\end{equation}
where $q$ is the distance that the ray travels from $\mathbf{o}$ in direction $\mathbf{d}$, $\mathbf{c}(\mathbf{r}(q), \mathbf{d})$ represents the color contribution (R, G, and B) at point $\mathbf{r}(q)$ along the ray for a specific direction $\mathbf{d}$, $\sigma(\mathbf{r}(q))$ is the occupancy of the scene at point $\mathbf{r}$ along the ray. $q_n$ and $q_f$ denote the bounds of the volume depth range and the accumulated opacity. $T(q)$ indicates the accumulated transmittance along the ray $\mathbf{r}$, which is further calculated by
\begin{equation}
\label{Tq}
    T(q) = e^{\left(-\int_{q_n}^{q}\sigma(\textbf{r}(l)) \right)} dl,
\end{equation}
where $l$ is the distance that the ray has traveled from its starting point $q_n$ to the current point $q$ along the ray's path. 
\begin{figure}
            \centering
            \includegraphics[width=0.49\textwidth]{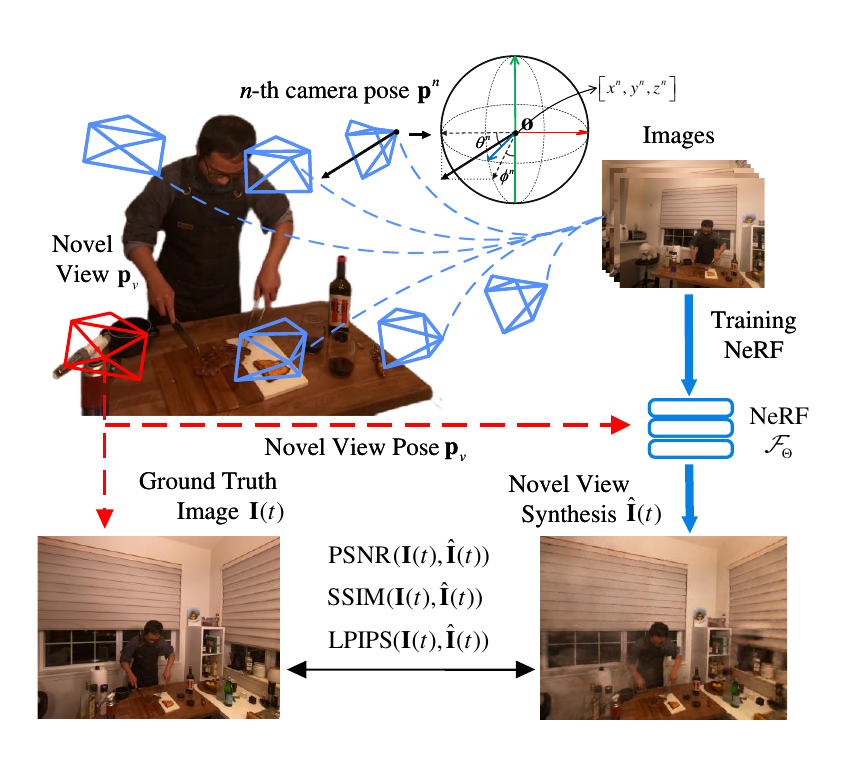}
           \caption{3D scene representation.}
           \label{fig:nerffig}
\end{figure}

\subsection{Performance Metrics}
To evaluate the performance of 3D scene representations, we employ the novel view synthesis approach~\cite{fridovich2023k, wang2020unsupervised, dynerf, nerf}. As shown in Fig.~\ref{fig:nerffig}, the novel view synthesis refers to the process of rendering an image from the view that was not originally captured or observed. By comparing different similarity metrics between the image of the reconstructed image and the ground truth, we evaluate the performance of novel view synthesis and the performance of 3D scene representations from different perspectives~\cite{tewari2022advances}. 

To quantify the performance of novel view synthesis, we adopt three widely used metrics, i.e., \gls{psnr}, \gls{ssim}~\cite{ssim}, and \gls{lpips}~\cite{lpips}. 
 
In the $t$-th time slot, the \gls{psnr} metric evaluates the per-pixel level picture fidelity and accuracy by comparing the ground truth of the image $\mathbf{I}(t)$ and the synthesized image $\mathbf{\hat{I}}(t)$ from 3D scene representations, 

\begin{equation}
    \text{PSNR}\left(\mathbf{I}(t), \mathbf{\hat{I}}(t)\right) = 10 \cdot \log_{10}\left(\frac{R_I^2}{\text{MSE}(\mathbf{I}(t), \mathbf{\hat{I}}(t))}\right),
    \label{psnr1}
\end{equation}
\begin{equation}
        \text{MSE}\left(\mathbf{I}(t), \mathbf{\hat{I}}(t)\right) = \frac{1}{L_H L_W}\sum_{m=1}^{L_H}\sum_{n=1}^{L_W}\left(i_{m, n}(t)-\hat{i}_{m, n}(t)\right)^2, 
\end{equation}
where $\mathbf{I}(t)$ is the image of the ground truth of the novel view, and $\mathbf{\hat{I}}(t)$ is the synthesized image from 3D scene representations. $i_{m, n}(t)$ and ${\hat{i}}_{m, n}(t)$ are the $m$-th row by $n$-th column pixel value of the given image $\mathbf{I}(t)$ and $\mathbf{\hat{I}}(t)$, respectively. The height and width pixel numbers of the images are denoted by $L_H$ and $L_W$, respectively, and $R_I = 2^{\kappa}-1$ is the maximum fluctuation for an image of $\kappa$-bit color per pixel. 

To evaluate the difference in luminance, contrast, and structural information, we utilize \gls{ssim}, 
\begin{equation}
\begin{aligned}
       &\text{SSIM}\left(\mathbf{I}(t), \mathbf{\hat{I}}(t)\right)
       \\
      &\quad\; =\frac{\left(2\mu_\mathbf{I}\mu_\mathbf{\hat{I}}+(k_1L_d)^2\right) \left((2\sigma_{c}+(k_2L_d)^2\right) }{\left(\mu_\mathbf{I}^2+\mu_\mathbf{\hat{I}}^2+(k_1L_d)^2\right)\left(\sigma_\mathbf{I}^2+\sigma_\mathbf{\hat{I}}^2+(k_2L_d)^2\right)},
\end{aligned}
\end{equation}
where the average pixel value of $\mathbf{I}(t)$ and $\mathbf{\hat{I}}(t)$ are denoted by $\mu_\mathbf{I}$ and $\mu_\mathbf{\hat{I}}$, respectively, the standard deviations of the pixel values of $\mathbf{I}(t)$ and $\mathbf{\hat{I}}(t)$ are denoted by $\sigma_\mathbf{I}$ and $\sigma_\mathbf{\hat{I}}$, respectively, the covariance between $\mathbf{I}(t)$ and $\mathbf{\hat{I}}(t)$ is $\sigma_{c}$, $k_1$ and $k_2$ are the parameters to avoid instability, and the dynamic range of the images is denoted by $L_d$. 

To evaluate the perceptional similarity between images, a learning-based perceptual image patch similarity metric, \gls{lpips} is also used. The idea is to estimate the visual similarity of human perception by learning a neural network model. The model uses a  convolutional neural network (CNN) to perform feature extraction on local patches of an image and calculates the similarity score between patches~\cite{lpips}. In the $t$-th time slot, the \gls{lpips} is
\begin{equation}
    \text{LPIPS}\left(\mathbf{I}(t), \mathbf{\hat{I}}(t)\right) = \mathcal{G}\left(d\left(\mathbf{I}(t), \mathbf{\hat{I}}(t)\right)\right),
    \label{lpips1}
\end{equation}
\begin{equation}
    d\left(\mathbf{I}(t), \mathbf{\hat{I}}(t)\right) = \sum^{L}_{l=1} \frac{1}{L_H L_W} {\lVert w_l \odot \left(\mathbf{I}'_l(t) - \mathbf{\hat{I}}'_l(t)\right) \rVert}_2^2,
    \label{lpips2}
\end{equation}
where $\mathbf{I}'_l(t)$ and $\mathbf{\hat{I}}'_l(t)$ are the features extracted at the $l$-th layer of the neural network with input $\mathbf{I}(t)$ and $\mathbf{\hat{I}}(t)$, respectively. The activation functions for each layer of the neural network are denoted by vector $w_l$. $\mathcal{G}$ is the neural network that predicts the \gls{lpips} metric value from input distance $d(\mathbf{I},\mathbf{\hat{I}})$. ${\lVert \cdot \rVert}_2$ is the $\ell_2$ norm distance and $\odot$ denotes the tensor product operation.

\section{Problem Formulation}
To achieve optimal fidelity 3D scene representation results, we proposed a \gls{drl} method that leverages the \gls{aoi} to decide if the received image needs to be involved in 3D scene representations and rendering to optimize the performance of novel view synthesis. Specifically, we propose to use the \gls{ppo} algorithm as the baseline method for its simplicity, effectiveness, and high sample efficiency~\cite{ppo}. Since our framework encompasses only the randomness induced by the one-step dynamics of the environment, we modify the standard \gls{ppo} algorithm to one single-step \gls{ppo} to solve the problem~\cite{singlestep}.

\subsection{State}
The state in the $t$-th time slot is the AoIs of the $N$ cameras, which is donated by $\mathbf{s}_t = [\Delta^1(t), \Delta^2(t),..., \Delta^N(t)] \in \mathbb{R}^{1\times N}$.

\subsection{Action}
The action to be taken in the $t$-th time slot is set to the decision of whether the $i$-th update of the $n$-th camera ${\tau}_i^n$ is used for 3D scene representations in the $t$-th time slot, $n=1,2,...,N$, i.e., $\mathbf{a}_t=\Omega(t)$.

\subsection{Reward}
Given the state $\mathbf{s}_t$ and action $\mathbf{a}_t$ in the $t$-th time slot, the instantaneous reward in the $t$-th time slot, is defined as 
\begin{equation}
\begin{aligned}
    r(\mathbf{s}_t, \mathbf{a}_t)  = w_1 \times \text{PSNR}(&\mathbf{I}(t), \mathbf{\hat{I}}(t)) \\
    + w_2 \times \text{SSIM}(\mathbf{I}(t), \mathbf{\hat{I}}&(t))  + w_3 \times \text{LPIPS}(\mathbf{I}(t), \mathbf{\hat{I}}(t)),
\end{aligned}
\end{equation}
which is a weighted sum of the three metrics we introduced in the last section. Depending on the application scenario, we can set $w_1$, $w_2$ and $w_3$ to be to different values.

\begin{algorithm}[t]
\caption{PPO}
\label{algorithm1}
\begin{algorithmic}[1]
\State \text{Input}: Initialize the parameters of channel model with communication delay distribution parameter $\lambda$, initial parameters of neural network $\theta_0$, \gls{nerf} novel view pose $\mathbf{p}_v$, training steps $T_t$, parameters of \gls{nerf} neural network $\alpha_\Theta$. 
\For{$t = 1, 2, ... T_t$}
    \State $\mathbf{s}_t \gets [\Delta^1(t), \Delta^2(t),..., \Delta^N(t)]$ Obtain the state from 
    \Statex \quad \; AoIs.
    \State $\mathbf{a}_t \gets \pi_{\theta_t}(\mathbf{s}_t)$ Take the action $\mathbf{a}_t$  from $\pi_{\theta_t}(\mathbf{s}_t)$.
    \State $\mathbf{\hat{I}}(t) \gets$ Train \gls{nerf} and render the image from novel
    \Statex \quad \; view $\mathbf{p}_v$.
    \State $r(\mathbf{s}_t, \mathbf{a}_t) \gets \text{LPIPS}(\mathbf{I}(t), \mathbf{\hat{I}}(t)), \text{SSIM}(\mathbf{I}(t), \mathbf{\hat{I}}(t)),$ 
    \Statex \quad \; $\text{PSNR}(\mathbf{I}(t), \mathbf{\hat{I}}(t))$~ Calculate the instantaneous reward 
    \Statex \quad \; by (\ref{psnr1})-(\ref{lpips2}).
    \State $A^{\pi_\theta} \gets \; Q^{\pi_{\theta_t}} (\textbf{s}_t, \textbf{a}_t) -  V^{\pi_{\theta_t}}(\textbf{s}_t)$~Calculate advantage 
    \Statex \quad \; with~(\ref{qvalue}), (\ref{value}).
    \State $\pi_{\theta_{t+1}} \gets A^{\pi_\theta}$ Take one-step policy update towards 
    \Statex \quad \; maximizing $\mathcal{L}(\textbf{s}_t, \textbf{a}_t, \theta_t, \theta)$ in~(\ref{loss}).
\EndFor
\State \textbf{Output}: Optimal policy $\pi^*_{\theta}$.
\end{algorithmic}
\end{algorithm}

\subsection{Problem Formulation}
The policy $\pi_\theta$ maps the state ${\mathbf{s}}_t$ to the distribution of ${{\mathbf{a}}_{t}}$, denoted by $\bm{\rho}_t \in \mathbb{R}^{2 \times N}$, defined as
\begin{align}\label{eq:sampling}
{\bm\rho_t^n} \triangleq
\begin{pmatrix}
   \Pr\{a_t^n=1\}\\
   \Pr\{a_t^n=0\}
\end{pmatrix}.
\end{align}
The policy is represented by a neural network denoted by $\pi_{\theta}(\mathbf{s}_t)$
, where $\theta$ are the training parameters.
Following the $\pi_\theta$ policy, the long-term reward is given by
\begin{equation}
    R^{\pi_\theta} = \mathbb{E}[\sum_{t=0}^{\infty} \gamma^t r(\mathbf{s}_t, \mathbf{a}_t)],
\end{equation}
where $\gamma$ is the reward discounting factor. To find an optimal policy $\pi_\theta^*$ that maximizes the long-term reward $R^{\pi_\theta}$, the problem is formulated as
\begin{align}
    &{\pi_{\theta}^{*}}  =  \mathop {\max }\limits_{\theta} Q^{\pi_{\theta}}({\bf{s}}_t, {\bf{a}}_t)\hfill,    \\
    \label{qvalue}
    Q^{\pi_{\theta}}({\bf{s}_t}, {\bf{a}_t}) = & \;{\mathop{\mathbb{E}}}[\sum_{t = 0}^\infty  {{ \gamma ^t}}r({\bf{s}}_t, {\bf{a}}_t) \mid {\bf{s}}_0={\bf{s}},\ {\bf{a}}_0={\bf{a}},\ \pi_{\theta}],
\end{align}
where $Q^{\pi_\theta}$ is the state-action value function.

\begin{figure*}
  \centering
  \begin{minipage}[b]{0.33\linewidth}
    \centering
    \subfigure[]{
    \label{PSNR}
    \includegraphics[scale=0.28]{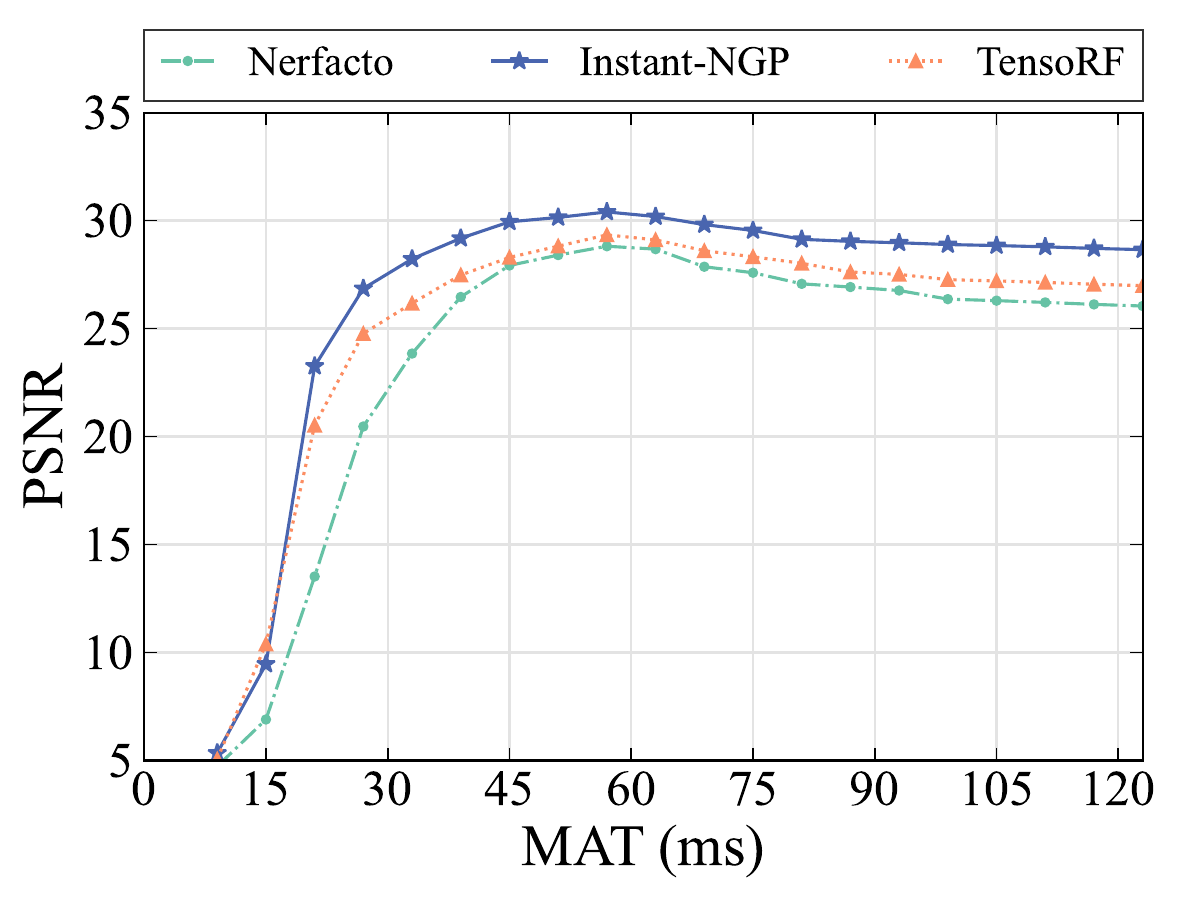}
    }
  \end{minipage}\hfill
  \begin{minipage}[b]{0.33\linewidth}
    \centering
    \subfigure[]{
    \label{SSIM}
    \includegraphics[scale=0.28]{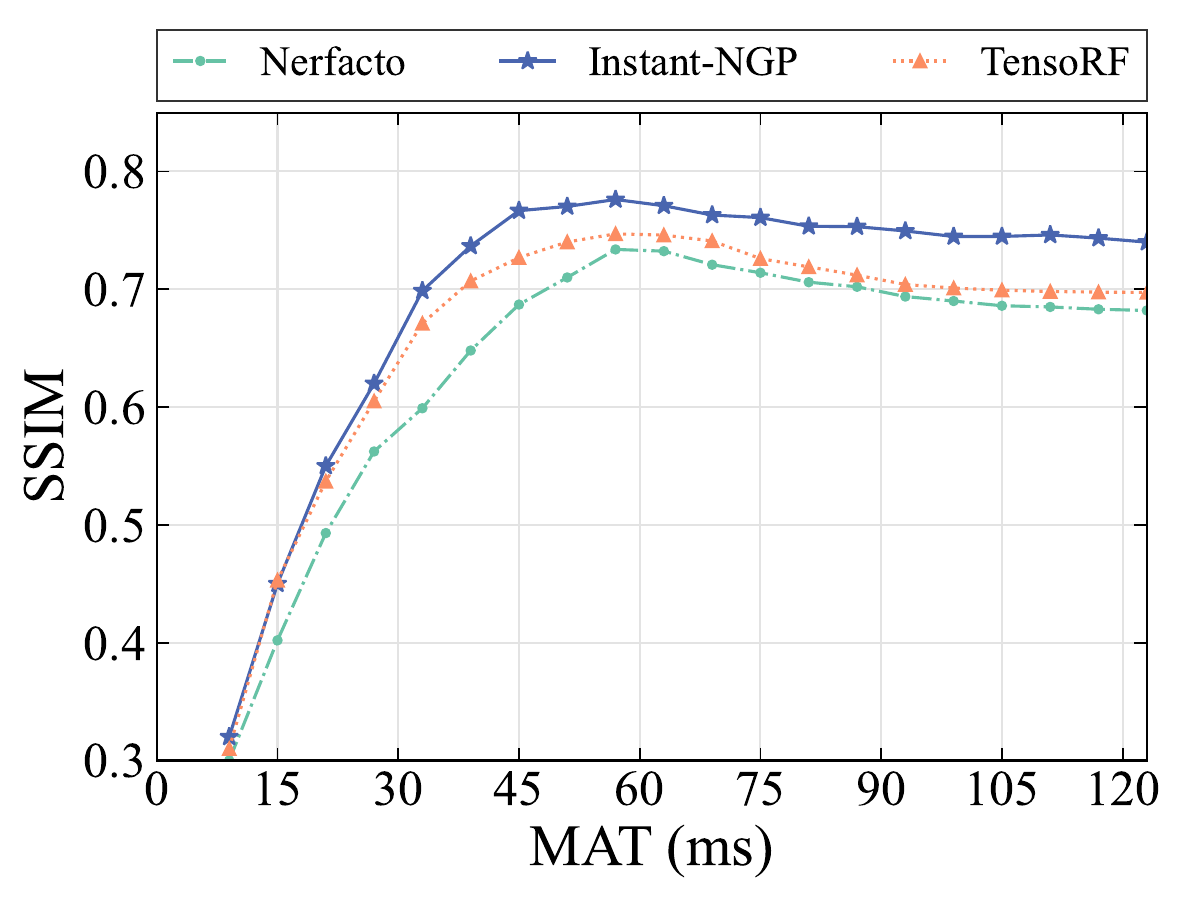}
    }
  \end{minipage}\hfill
  \begin{minipage}[b]{0.33\linewidth}
    \centering
    \subfigure[]{
    \label{LPIPS}
    \includegraphics[scale=0.28]{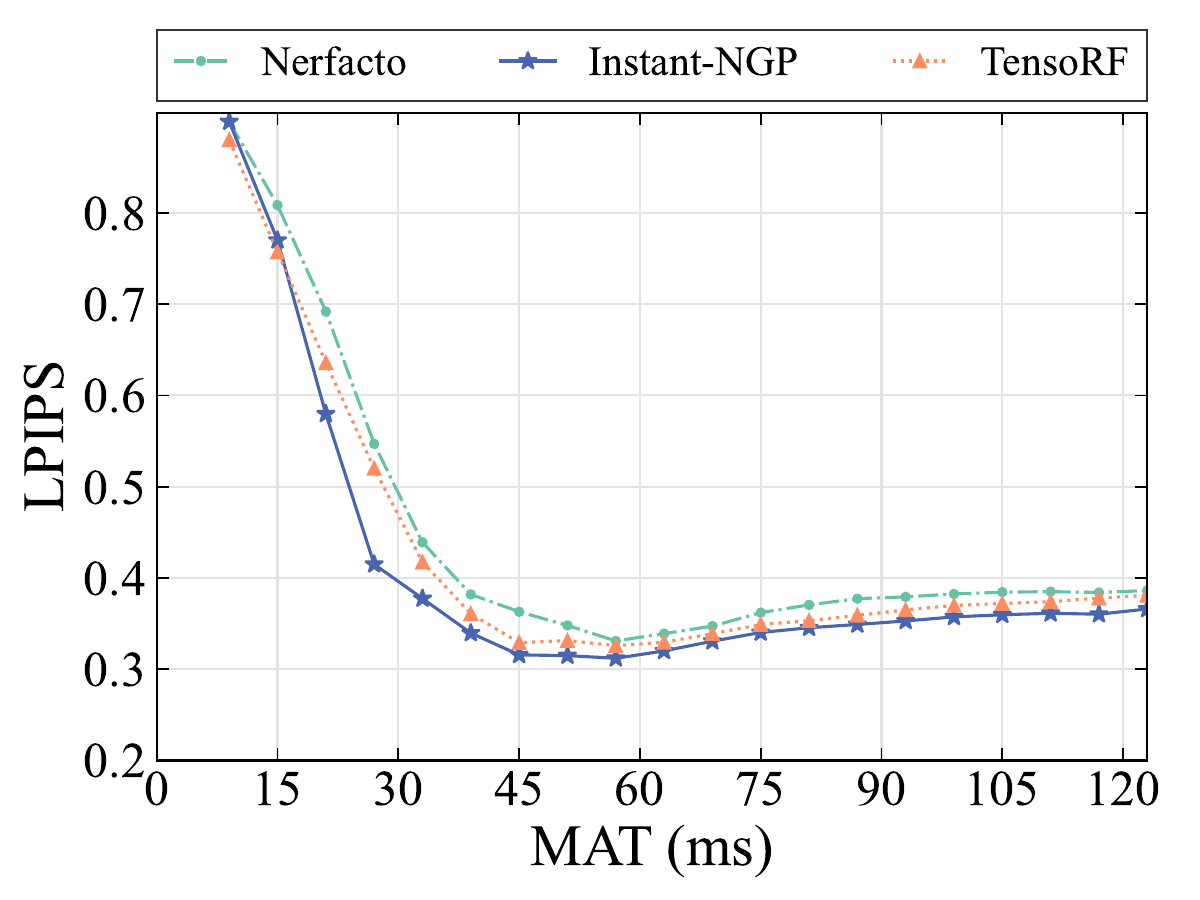}
    }
  \end{minipage}
  \caption{Timeliness-fidelity tradeoff in three performance metrics, i.e., PSNR, SSIM, and LPIPS.} 
  \label{fig:tradeoff} 
\end{figure*}
\section{Single-Step PPO Algorithm}
The \gls{ppo} algorithm updates the parameters of the policy neural network $\theta_t$ by
\begin{align}\label{loss}
\mathcal{L}({\mathbf{s}}_t,{\mathbf{a}}_t &,\theta_t,\theta) = \min \left(\frac{\pi_{\theta}({\mathbf{a}}_t \mid {\mathbf{s}}_t)}{\pi_{\theta_{t}}({\mathbf{a}}_t \mid {\mathbf{s}}_t)}{A^{\pi_{\theta_t}}({\mathbf{s}}_t,{\mathbf{a}}_t)},\right. \\ \notag
&\left.\text{clip}\left(\frac{{\pi_\theta}({\bf{a}}_t \mid {\bf{s}}_t)}{\pi_{\theta_t}({\bf{a}}_t \mid {\bf{s}}_t)}, 1-\epsilon, 1+\epsilon \right)A^{\pi_{\theta_t}}({\bf{s}}_t,{\bf{a}}_t)\right),
\end{align} 
where $A^{\pi_{\theta_t}}$ is the advantage function estimating the advantage of taking action $\bf{a}_t$ in state $\bf{s}_t$~\cite{gae}, which is expressed by
\begin{align}
\label{advantage}
A^{\pi_{\theta_t}} = &\; Q^{\pi_{\theta_t}}(\textbf{s}_t, \textbf{a}_t) -  V^{\pi_{\theta_t}}(\textbf{s}_t),\\
\label{value}
V^{\pi_{\theta_t}}({\bf{s}}_t) = &\; {\mathop{\mathbb{E}}}[\sum_{t = 0}^\infty  {{ \gamma^t}}r({\bf{s}}_t, {\bf{a}}_t) \mid {\bf{s}}_0={\bf{s}},\ \pi_{\theta}], 
\end{align}
where $V^{\pi_{\theta_t}}$ is the state-value function.

The details of the proposed single-step \gls{ppo} algorithm are shown in~\cref{algorithm1}. First, different parameters of neural networks and channel models are initialized. In the $t$-th time slot, based on the \glspl{aoi} of received images at the edge server, we obtain the state, $\mathbf{s}_t = [\Delta^1(t), \Delta^2(t),..., \Delta^N(t)]$ and take the action, $\mathbf{a}_t$. Then, by calculating the weighted sum reward from the \gls{nerf} and rendering results, we obtain the instantaneous reward. After that, we estimate the advantage $A(\mathbf{s}_t,\mathbf{a}_t)$ of taking action $\mathbf{a}_t$ in state $\mathbf{s}_t$ with the state-action value function (\ref{qvalue}) and state-value function (\ref{value}). With the advantage $A(\mathbf{s}_t,\mathbf{a}_t)$ we take one step gradient descend to update the current policy $\pi_{\theta_t}$ to $\pi_{\theta_{t+1}}$ by maximizing the \gls{ppo} loss function (\ref{loss}).

\section{Simulation Setup}
To evaluate the result of the 3D scene representations, we use the DyNeRF dataset as it is widely used in the 3D scene representations and novel view synthesis~\cite{dynerf}. Specifically, 10-second 30-FPS multi-view videos recorded by 19 cameras from different view angles are provided in the dataset.
\begin{figure*}
            \centering
            \includegraphics[width=0.87\textwidth]{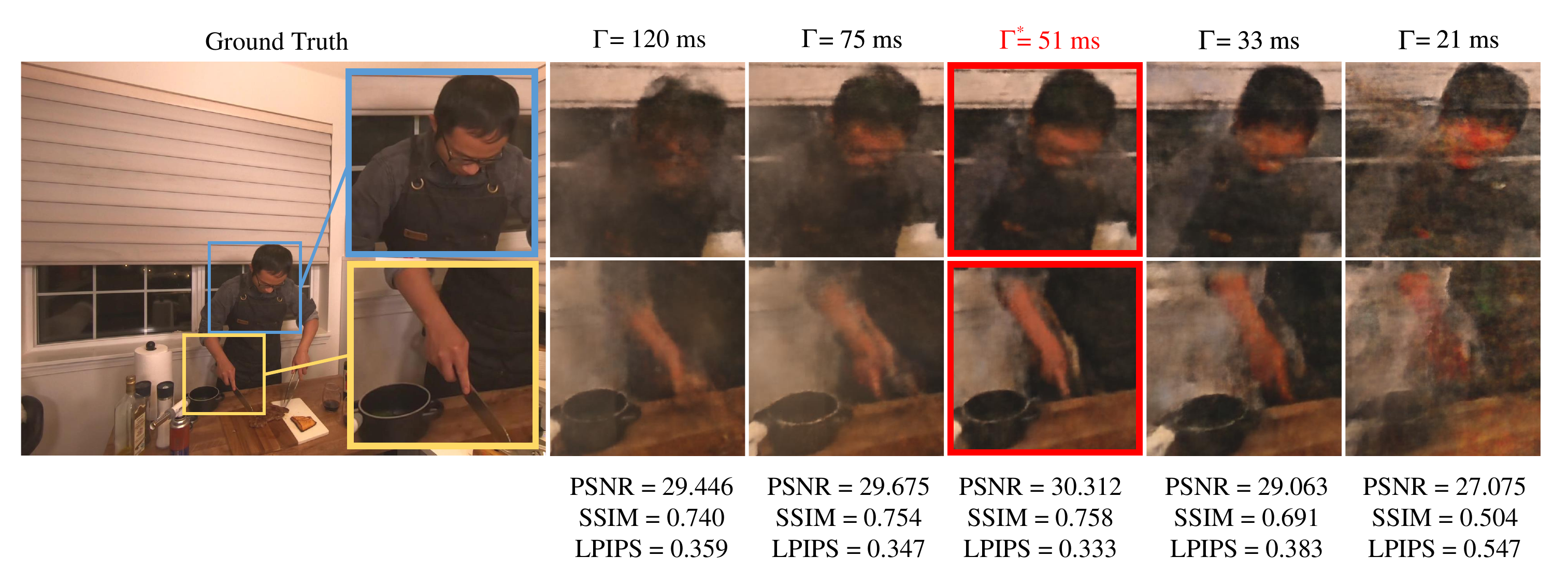}
           \caption{Novel view synthesis results with different MATs.}
           \label{fig:results1}
\end{figure*}
For comparison, we take 18 of 19 videos for training the 3D scene representations, i.e. $N=18$, and the remaining one is set as the ground truth for testing the novel view synthesis. 
 
The generating interval of each camera $C$ is set to 30~ms.
The transmission delay $Y^n_i$ for each camera follows an exponential distribution with an average delay of 60 ms.
The length of each time slot is set to $T_s = 1$ ms.
The reward discounting factor $\gamma$ in the single-step \gls{ppo} algorithm is set to 1. 

To investigate the tradeoff between timeliness and fidelity in real-time 3D scene representations, we propose to use three well-known \gls{nerf} methods:

\begin{figure}
    \centering
     \includegraphics[width=0.7\linewidth]{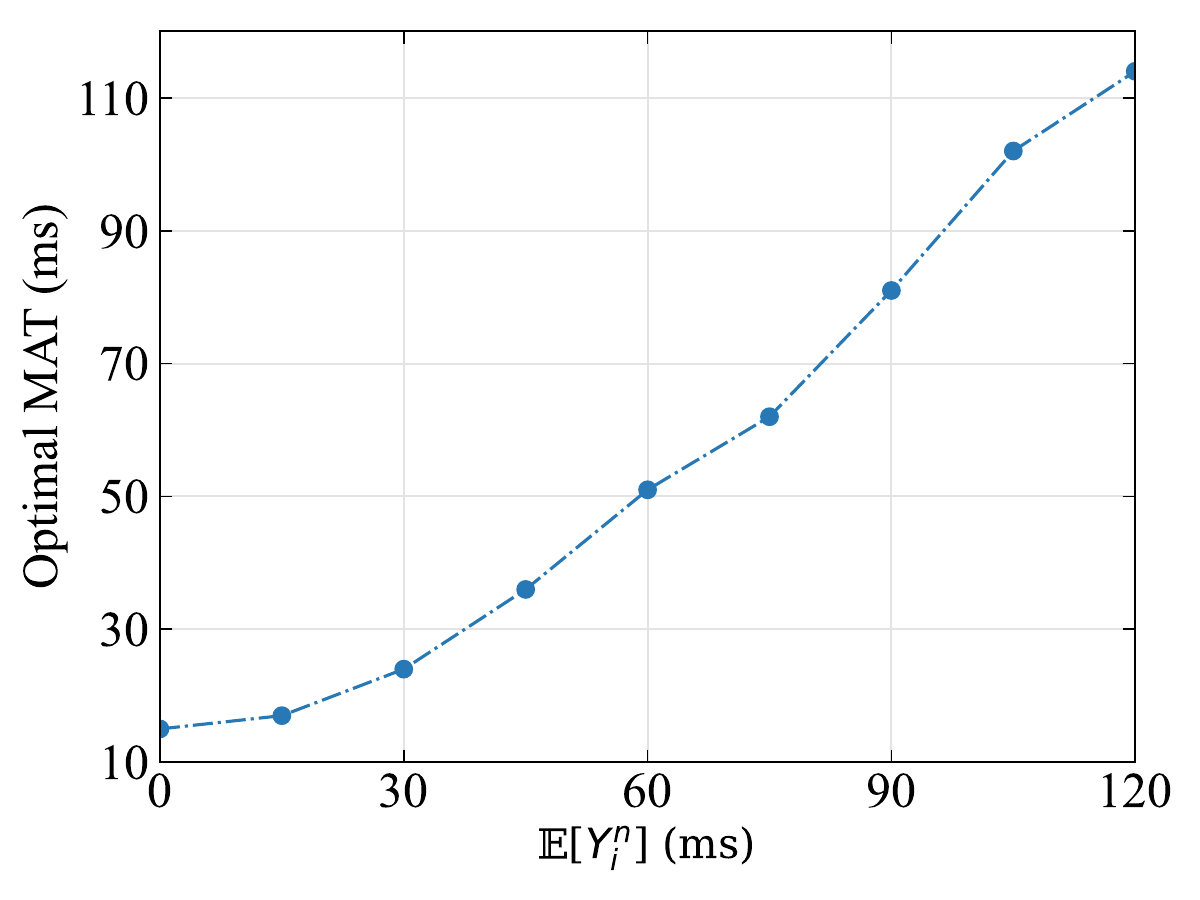}
      \caption{The relationship between optimal MAT $\Gamma$ and expected value of transmission duration $\mathbb{E}{[Y_i^n]}$.}
     \label{lambda}
\end{figure}
\begin{figure}
    \centering
     \includegraphics[width=0.7\linewidth]{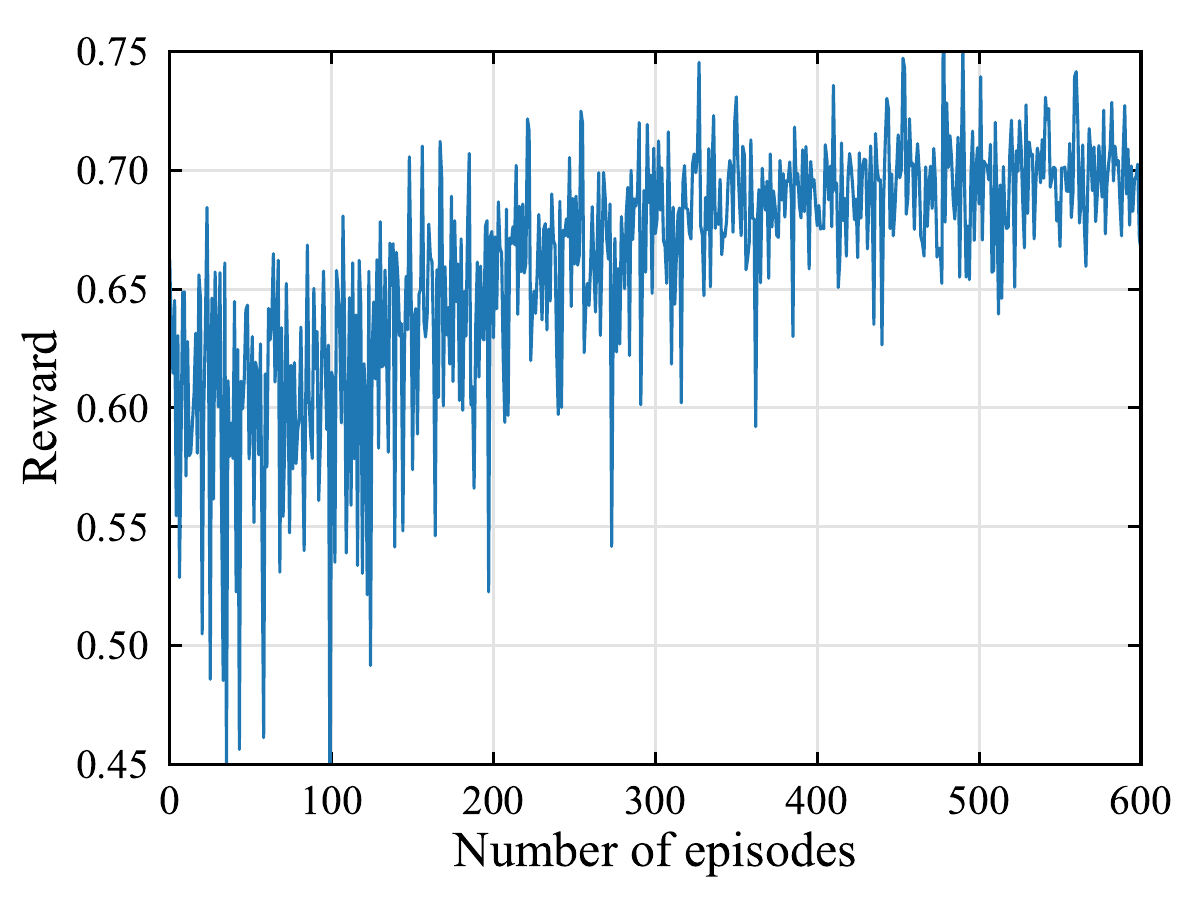}
      \caption{Reinforcement learning reward in each training episode.}
     \label{fig:reward}
\end{figure}

\subsubsection{\gls{ngp}~\normalfont{\cite{instantngp}}}
\gls{ngp} is a \gls{nerf}-based 3D scene representation method. Utilizing a C++ embedded neural network structure, hash coding, and \gls{cuda}, \gls{ngp} achieves state-of-the-art training speed among all \gls{nerf} methods.

\subsubsection{TensoRF \normalfont{\cite{tensorf}}}
TensoRF is also a \gls{nerf}-based 3D scene representation method. It replaced the fully connected neural network layers in \gls{nerf} with a 4D tensor to represent the volume field of the scene. This approach factorized the feature grid into multiple compact low-rank tensor components to achieve efficient training both in computing memory and time.

\subsubsection{Nerfacto \normalfont{\cite{nerfstudio}}}
Nerfacto integrates camera pose refinement and per-image appearance conditioning to augment reconstruction quality. It applies the hash coding from \gls{ngp} to accelerate training. Compared with \gls{ngp}'s \gls{cuda}-based core computing module, Nerfstudio is programmed in Python and thus needs more computation time.

\section{Performance Evaluation}
In~\ref{timeliness}, to show the timeliness-fidelity tradeoff, we consider a threshold method. In~\ref{rl}, to show the optimal performance achieved by the proposed single-step PPO algorithm, we evaluate the training and testing results. 

\subsection{Timeliness-Fidelity Tradeoff with a threshold-based Method}\label{timeliness}
To investigate the timeliness-fidelity tradeoff, we compare the training performance with different \glspl{mat} denoted by $\Gamma$. Specifically, in the $t$-th time slot, if $\Delta^n(t) < \Gamma \in [0,120]$~ms, $\widetilde{\tau}^n(t)$ is used for 3D scene representations. Otherwise, $\widetilde{\tau}^n(t)$ will not be used. We evaluate the performance metrics by computing the average performance in 40 simulations.
Figure~\ref{fig:tradeoff} illustrates three performance metrics of novel view synthesis, i.e., PSNR, SSIM, and LPIPS vs. different \glspl{mat}. The agent only uses the images with \glspl{aoi} lower than the \gls{mat} for 3D scene representations. The result shows that there exists a tradeoff between the performance of 3D scene representations and the \gls{mat}.

Figure~\ref{fig:results1} shows a certain novel view synthesis with different \glspl{mat}. As expected, the novel view synthesis achieves the best performance from the naked eye at about $\Gamma$ = 51~ms. If the \gls{mat} is too large (such as 120~ms), the fidelity of the image is high, but it is delayed compared with the ground truth. If the \gls{mat} is too small, the image is synchronized with the ground truth, but the fidelity is poor. By optimizing the set for 3D scene representations $\mathbb{D}(t)$, we can obtain a good tradeoff between timeliness and fidelity that maximizes the image quality.

Figure~\ref{lambda} demonstrates the relationship between the optimal \gls{mat} and the expected value of transmission delay $\mathbb{E}[Y_i^n]$. The results show that the optimal \gls{mat} exhibits a monotonic increasing trend to $\mathbb{E}[Y_i^n]$. In scenarios with poor channel conditions, it is better to increase the MAT.

\subsection{3D Scene Representations with Single-step PPO}\label{rl}
In Fig.~\ref{fig:reward}, we evaluate the reward achieved by the single-step \gls{ppo}. The parameters are set to $w_1=0.02, w_2=0.5, w_3=-1$. With \gls{ngp} as the \gls{nerf} module, we train the algorithm for 600 episodes to show the training performance.

The results show that our algorithm converges after 400 training episodes. In the training stage, the optimal policy $\pi_\theta^*$ can achieve an average \gls{psnr} = 30.54, \gls{ssim} = 0.775, \gls{lpips} = 0.342 and in testing average \gls{psnr} = 30.12, \gls{ssim} = 0.779, \gls{lpips} = 0.350.

\section{Conclusion}
In this paper, we established a framework for investigating the tradeoff between timeliness and fidelity in real-time 3D scene representations. We evaluated the effects of communication delay on the 3D scene representations by adjusting different \glspl{mat}. Different well-known 3D scene representation methods and the weighted sum metric were used for simulations. The results indicated that there existed a tradeoff between timeliness and fidelity which was significantly affected by the communication delay, and the set of images for 3D scene representations was optimized by our proposed single-step \gls{ppo} algorithm.

\bibliographystyle{IEEEtran}
\bibliography{bib}

\vspace{12pt}


\end{document}